\documentclass[10pt,twocolumn,letterpaper]{article}

\usepackage[pagenumbers]{cvpr} %

\usepackage{graphicx}
\usepackage{amsmath}
\usepackage{amssymb}
\usepackage{booktabs}
\usepackage{makecell}
\usepackage[export]{adjustbox}
\usepackage[normalem]{ulem}
\usepackage{textpos}  %
\usepackage[misc]{ifsym}
\usepackage{animate}

\usepackage{tabulary,multirow,overpic,xcolor,subfloat}
\usepackage[pagebackref=false, breaklinks=true, letterpaper=true, colorlinks,
            citecolor=citecolor, linkcolor=linkcolor, bookmarks=false]{hyperref}
\definecolor{citecolor}{HTML}{0071BC}
\definecolor{linkcolor}{HTML}{ED1C24}

\newcommand{\app}{\raise.17ex\hbox{$\scriptstyle\sim$}}

\newcolumntype{x}[1]{>{\centering\arraybackslash}p{#1pt}}
\newcolumntype{y}[1]{>{\raggedright\arraybackslash}p{#1pt}}

\newlength\savewidth

\makeatletter\renewcommand\paragraph{\@startsection{paragraph}{4}{\z@}
  {.5em \@plus1ex \@minus.2ex}{-.5em}{\normalfont\normalsize\bfseries}}\makeatother

\newcommand\blfootnote[1]{\begingroup\renewcommand\thefootnote{}\footnote{#1}\addtocounter{footnote}{-1}\endgroup}

\DeclareMathAlphabet\mathbfcal{OMS}{cmsy}{b}{n}

\definecolor{Gray}{gray}{0.5}

\newcommand{\modelname}{MagicEdit\xspace}

\usepackage[capitalize]{cleveref}
\crefname{section}{Sec.}{Secs.}
\Crefname{section}{Section}{Sections}
\Crefname{table}{Table}{Tables}
\crefname{table}{Tab.}{Tabs.}

\def\cvprPaperID{*****}
\def\confName{CVPR}
\def\confYear{2022}

\crefname{section}{\S}{\S\S}
\crefname{subsection}{\S}{\S\S}
\crefformat{table}{Table~#2#1#3}
\crefformat{figure}{Figure~#2#1#3}
\crefformat{equation}{Equation~(#2#1#3)}
\crefformat{algorithm}{Algorithm~#2#1#3}
\crefformat{appendix}{Appendix~#2#1#3}

\newcommand{\authorskip}{\hspace{2.5mm}}

\usepackage{amsmath,amsfonts, mathtools, bm, bbm, nicefrac}

\newcommand{\E}{\mathbb{E}}

\DeclareMathOperator*{\argmin}{arg\,min}

\def\vc{{\bm{c}}}

\def\vs{{\bm{s}}}

\def\vx{{\bm{x}}}

\DeclareMathAlphabet{\mathsfit}{\encodingdefault}{\sfdefault}{m}{sl}
\SetMathAlphabet{\mathsfit}{bold}{\encodingdefault}{\sfdefault}{bx}{n}

\begin{document}

\title{\modelname:
High-Fidelity and Temporally Coherent Video Editing}

\author{
 Jun Hao Liew$^*$ \authorskip 
 Hanshu Yan$^*$ \authorskip
 Jianfeng Zhang \authorskip
 Zhongcong Xu \authorskip 
 Jiashi Feng \authorskip \\
 ByteDance Inc.\\
 {\small \url{https://magic-edit.github.io/}}
}

\twocolumn[{%
\renewcommand\twocolumn[1][]{#1}%
\maketitle
\begin{center}
    \centering
    \vspace{-1em}
    \animategraphics[width=\textwidth,loop]{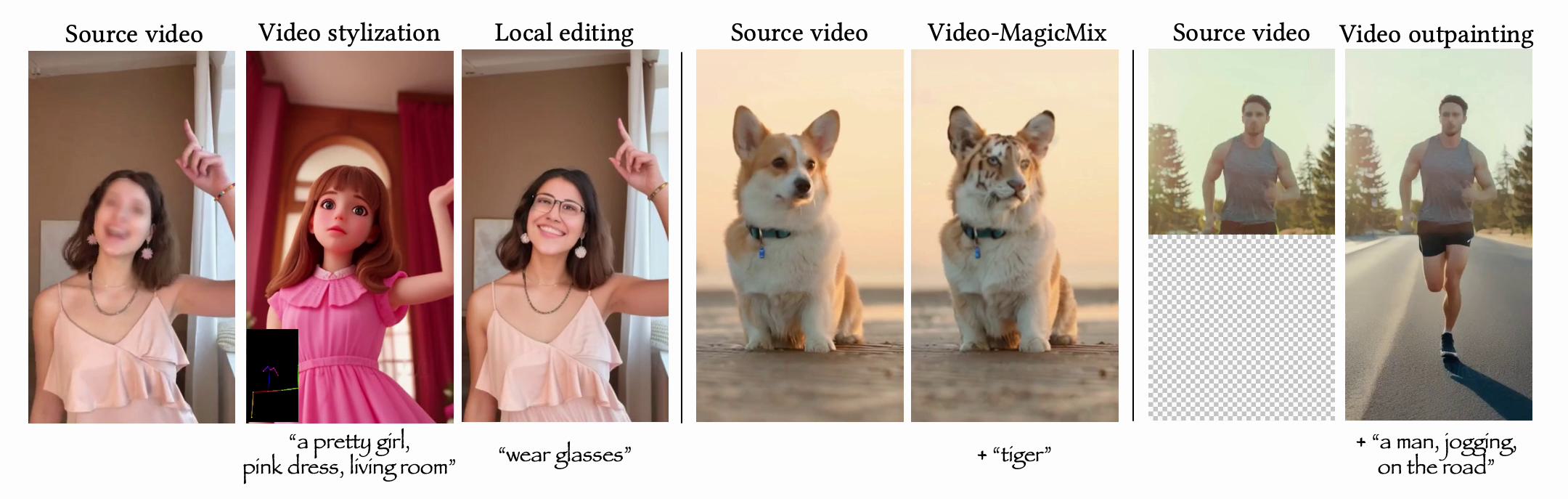}{figs/teaser/}{0}{15}
    \captionof{figure}{\textbf{\modelname} explicitly disentangles the learning of content, structure and motion signal to achieve high-fidelity and temporally coherent video editing. As a result, \modelname supports a variety of video editing applications, including video stylization, local editing, video-MagicMix (mixing of two concepts to create a novel concept) and video outpainting.
    Please note that this figure contains video clips. We encourage readers to \textcolor{magenta}{click and play} using Adobe Acrobat. \textbf{Faces in source videos are blurred} to protect identities.
    }
    \label{fig:teaser}
\end{center}%
}]
\blfootnote{$^*$Equal Contribution}

\begin{abstract}
In this report, we present \modelname, a surprisingly simple yet effective solution to the text-guided video editing task. 
We found that high-fidelity and temporally coherent video-to-video translation can be achieved by explicitly disentangling the learning of content, structure and motion signals during training.
This is in contradict to most existing methods which attempt to jointly model both the appearance and temporal representation within a single framework, which we argue, would lead to degradation in per-frame quality.
Despite its simplicity, we show that \modelname supports various downstream video editing tasks, including video stylization, local editing, video-MagicMix and video outpainting.
%
\end{abstract}

\section{Introduction}
Video editing plays an ubiquitous role in creating fascinating visual effects for films, short videos, \etc. However, professional editing is not only complex and time-consuming, but also challenging for novice users. As a result, there is an increasing demand for easy-to-use and performant video editing tools. 
Recently, we have witnessed a rapid development of video editing algorithms~\cite{esser2023structure,yang2023rerender,wu2022tune,wang2023zero,liu2023video,shin2023edit} thanks to the introduction of powerful text-conditioned diffusion models trained on large-scale datasets (\eg, DALL-E 2~\cite{ramesh2022hierarchical}, Imagen~\cite{saharia2022photorealistic}, Stable Diffusion~\cite{rombach2022high}).
In general, there are two ways to
extend image diffusion models for the video editing tasks: per-frame methods and per-clip methods.

Per-frame methods treat a video clip as a sequence of frames and run image editing on each frame independently. These methods often require some ad-hoc tricks to reduce temporal inconsistency, such as cross-frame attention~\cite{wu2022tune}, flow warping~\cite{yang2023rerender}, latents matching/ fusion~\cite{ceylan2023pix2video} \etc. 
However, these strategies can only maintain high-level styles and coarse shapes, and are less effective in preserving fine-grained details and texture across frames.
In addition, these methods often struggle when there exists large motion.

Per-clip methods, on the other hand, treat a video clip as a 3D spatio-temporal volume and directly edit the entire video. 
These methods typically inflate the image diffusion model into a video model by adding temporal layers. 
Among these, a popular line of research~\cite{wu2022tune,wang2023zero,liu2023video,shin2023edit} is to either fine-tune the pre-trained model on the input video to generate videos with similar motion, or utilize Null-text Inversion~\cite{mokady2023null} for video inversion.
However, since fine-tuning or optimization is needed for every input video, these methods suffer from low efficiency.
%
Gen-1~\cite{esser2023structure}, on the other hand, incorporates temporal-aware structures and learns motion priors from large-scale video datasets, demonstrating remarkable video editing performance.
In general, compared to per-frame methods, temporal inconsistency is typically less of an issue for per-clip methods due to the explicit modeling of motion signal. 
Nevertheless, since these methods update the whole networks, the domain knowledge of the original text-to-image model is inevitably hurt, resulting in degradation of per-frame quality.

In this report, we discover a surprisingly simple yet effective recipe for text-guided video editing, \ie, by explicitly disentangling the learning of content, structure and motion during training, we can easily achieve high-fidelity, temporally consistent video-to-video translation.
With this, we present \modelname, which supports a variety of downstream editing tasks, including video stylization, local editing and video-MagicMix~\cite{liew2022magicmix} and video outpainting.

\begin{figure}[t]
    \centering
    \includegraphics[width=0.45\textwidth]{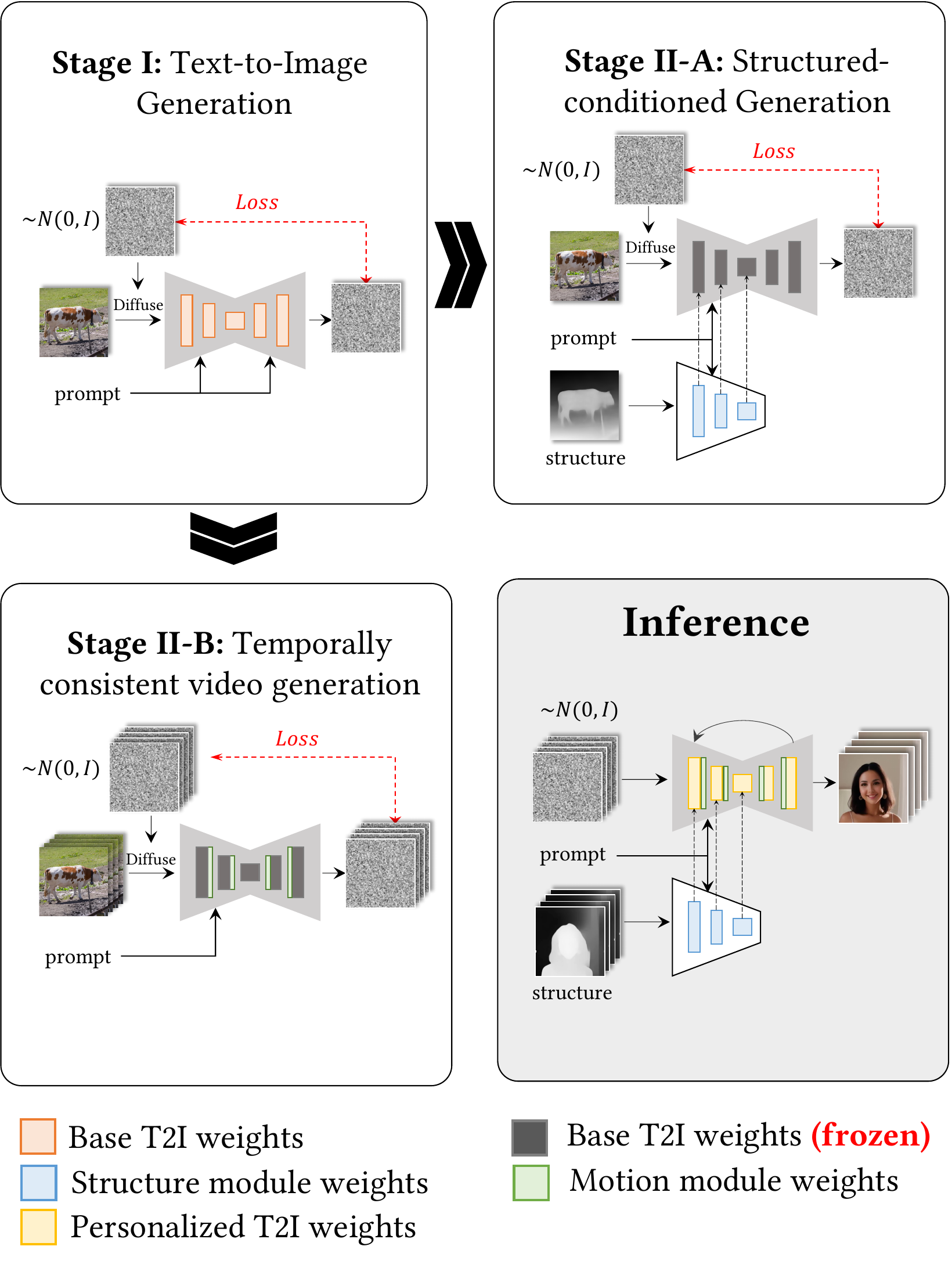}
    \caption{\textbf{The pipeline of \modelname. }Explicit disentanglement of content, structure and temporal smoothness during training is the key to high-fidelity temporally coherent video editing.}
    \label{fig:pipeline}
\end{figure}

\section{\modelname}
Given a video sequence of dimension $F \times H \times W \times 3$, where $F$ is the number of frames, $H, W$ are height and width, respectively, and a prompt description $\vc$ (\eg, ``{\tt a pretty girl, pink dress}" in Fig.~\ref{fig:teaser}), our goal is to edit the content of the video while preserving its structure. 
Specifically, we solve this task by learning a generative model $p(\vx|\vc,\vs)$ of videos $\vx=[\vx_1, \cdots, \vx_F]$, conditioned on text prompt $\vc$ and structure representation $\vs=[\vs_1, \cdots, \vs_F]$. We mathematically formulate this as:
\begin{equation*}
    \mathbf{\Theta} = \argmin_\Theta \E_{\vx, \vc, \vs} \left[ \sum_{\vx_i \in \mathbf{\vx}} \mathcal{L}_{\rm}(\vx_i, \vc, \vs, \theta_{\rm c}, \theta_{\rm s}, \theta_{\rm m})  \right]
\end{equation*}
where $\mathcal{L}$ refers to the noise estimation loss and $\mathbf{\Theta} = \{ \theta_{\rm c}$, $\theta_{\rm s}$, $\theta_{\rm m} \}$. In specific, $\theta_{\rm c}$ represents the UNet parameters of the text-to-image generation model; $\theta_{\rm s}$ refers to the parameters of the structure conditioning module; and $\theta_{\rm m}$ denotes the parameters of temporal/motion layers. We explicitly disentangle the modeling of content, structure and motion via stage-wise training as follows:

\medskip \noindent \textbf{Stage I: Text-to-Image generation. }
In the first stage, we train a base text-to-image (T2I) diffusion model to encourage each generated frame $\vx_i$ to adhere to the given text prompt $\vc$. 
\begin{equation*}
    \bar{\theta}_{\rm c} = \argmin_{\theta_{\rm c}} \E_{\vx_i, \vc} \mathcal{L}( \vx_i, \vc, \theta_{\rm c})
\end{equation*}
In this work, we choose {\tt stable-diffusion-v1-5} as our base T2I generation model.

\medskip \noindent \textbf{Stage II-A: Structure-conditioned generation. }
An important property of video editing is to ensure that each video frame follows the structure or trajectory of the source video (\eg, depth/ pose/ shape, \etc). For example, given a video of a girl swinging her arms (Fig.~\ref{fig:teaser}), each edited frame $\vx_i$ should follow the corresponding pose $\vs_i$ in the given video. 

To achieve this, we train a structure-conditioned module parameterized by $\theta_{\rm s}$ while \textbf{freezing} the pre-trained UNet parameters $\bar{\theta}_{\rm c}$. We follow the approach of ControlNet~\cite{zhang2023adding} for per-frame structure-preserving generation.
\begin{equation*}
    \bar{\theta}_{\rm s} = \argmin_{\theta_{\rm s}} \E_{\vx_i, \vs} \mathcal{L}(\vx_i, \vs, \theta_{\rm s};  \bar{\theta}_{\rm c})
\end{equation*}

\medskip \noindent \textbf{Stage II-B: Temporally consistent video generation. }
Lastly, to ensure the video frames remain temporally coherent, we train a motion module to enforce cross-frame consistency. For simplicity, we choose the vanilla temporal transformers from \cite{guo2023animatediff} as the design of our motion module. Once again, we train the motion module, parameterized by $\theta_{\rm m}$ while \textbf{freezing} the learned UNet parameters $\bar{\theta}_{\rm c}$. 
\begin{equation*}
    \bar{\theta}_{\rm m} = \argmin_{\theta_{\rm m}} \E_{\vx, \vc} \left[ \sum_{x_i \in \mathbf{x}} \mathcal{L}(\vx_i, \vc, \theta_{\rm m}; \bar{\theta}_c)  \right]
\end{equation*}

\medskip \noindent \textbf{Inference. }
During inference, we simply combine the three individually trained modules.
It is worth noting that, since the base T2I weights remain frozen when training the structure-conditioned and motion module, during inference, we can simply \textbf{swap the base T2I weights} ({\tt stable-diffusion-v1-5}) with different personalized Stable Diffusion models from CivitAI~\footnote{\url{https://civitai.com/}} for different styles and better appearance (Fig.~\ref{fig:pipeline} bottom right). 

\medskip \noindent \textbf{Implementation details. }
Vertical and horizontal videos are resized such that its shorter size is 320. Square videos are resized to 512$\times$512.
For each video clip, we sample 16 frames with fixed interval.
We use 25 step DDIM sampler and set the classifier-free guidance scale to 7.5.
Following \cite{guo2023animatediff}, we employ a linear beta schedule. 

\medskip \noindent \textbf{Discussion. }
We argue that the most important key to high-fidelity and temporally coherent video editing lies in the \textbf{explicit disentanglement of the three modules during training.}
More specifically, the text-to-image diffusion UNet should remain \textbf{frozen} when training the structure-following module and motion module. 
This is because video training data is of lower quality as compared to image data and often consists of motion blur.
In other words, jointly modeling all the three components, as done in most existing works, would lead to degraded per-frame quality.
To counter this, one needs to collect large-scale high quality video data, which is prohibitively expensive.

Note that, while none of these components/ modules are new, the main contribution of this work is to showcase that, explicitly disentangling the three sources of signal is the key towards high-quality temporally smooth video editing. 
We hope that this finding could shed lights on future video generation and editing research.

\section{Applications}
Next, we discuss the possible applications of \modelname, including stylization, local editing, video-MagicMix and video outpainting.

\begin{figure*}
    \centering
    \vspace{2em}
    \includegraphics[width=\textwidth]{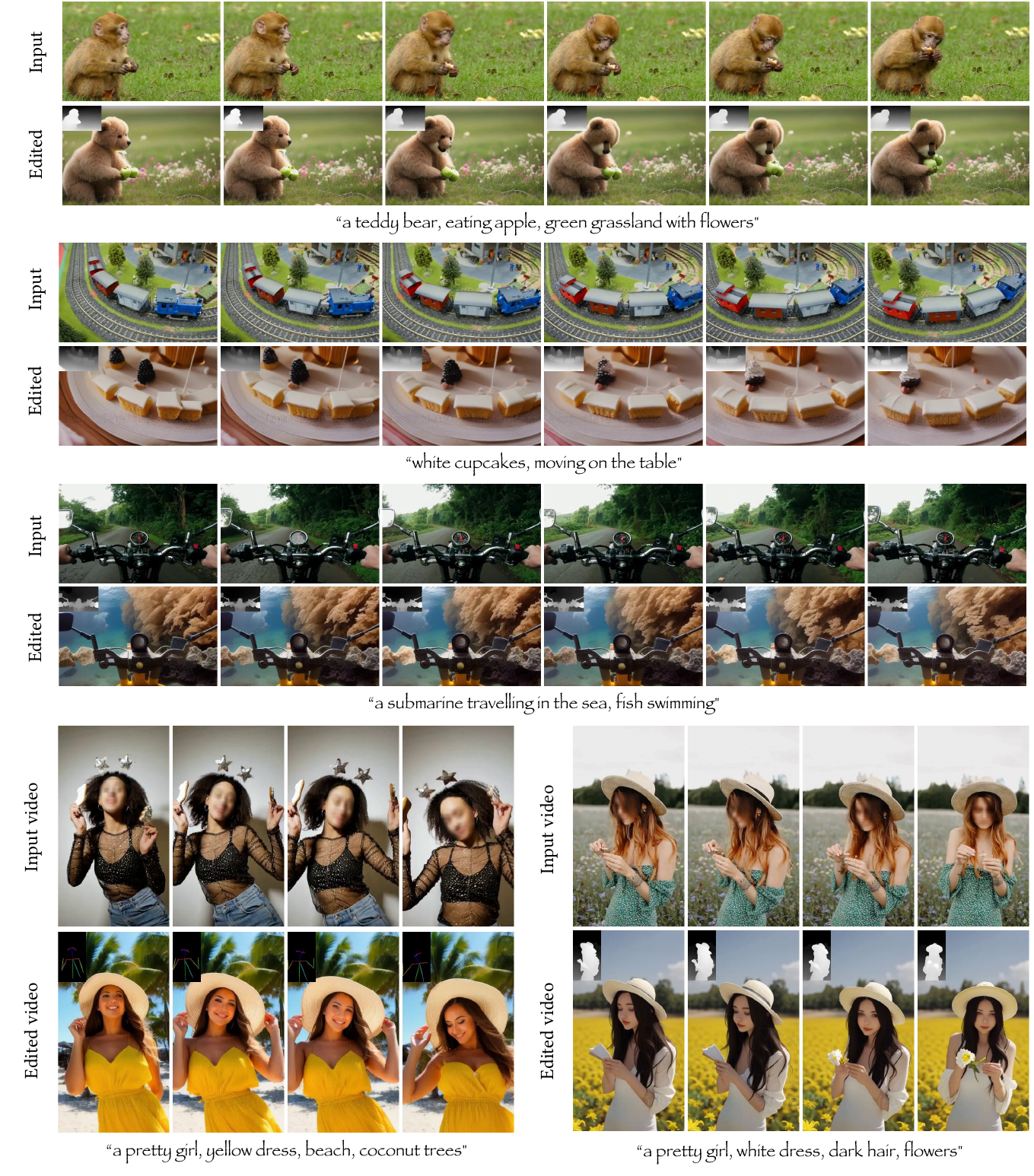}
    \caption{\textbf{Video stylization.} We generate new scenes with different subject(s) and different background while preserving the structure of original videos. The conditioned disparity maps and key points are shown in the top left of each edited video. \textbf{Faces in source videos are blurred} to protect identities.}
    \label{fig:stylization}
\end{figure*}

\medskip \noindent \textbf{Video stylization. } Video stylization enables one to (1) transform the source video into a new video with a style-of-interest (\eg, realistic, cartoon), or (2) creating a new scene with different subject (\eg, dog $\rightarrow$ cat) and different background (\eg, living room $\rightarrow$ beach).
Given a source video, we first extract its structure representation (\eg, we extract disparity maps with MiDaS~\cite{ranftl2020towards} or human pose with OpenPose~\cite{cao2017realtime}). Next, we swap base T2I weights with different personalized models from CivitAI (\eg, RealisticVision~\footnote{\url{https://civitai.com/models/4201}}, majicMix Realistic~\footnote{\url{https://civitai.com/models/43331}}, Disney Pixar Cartoon Type A~\footnote{\url{https://civitai.com/models/65203}}) for different styles. 
Following \cite{guo2023animatediff}, for each personalized model, we follow the prompts format provided at the model homepage. 
We show some examples in Fig.~\ref{fig:stylization}.

\begin{figure*}
    \centering
    \includegraphics[width=0.9\textwidth]{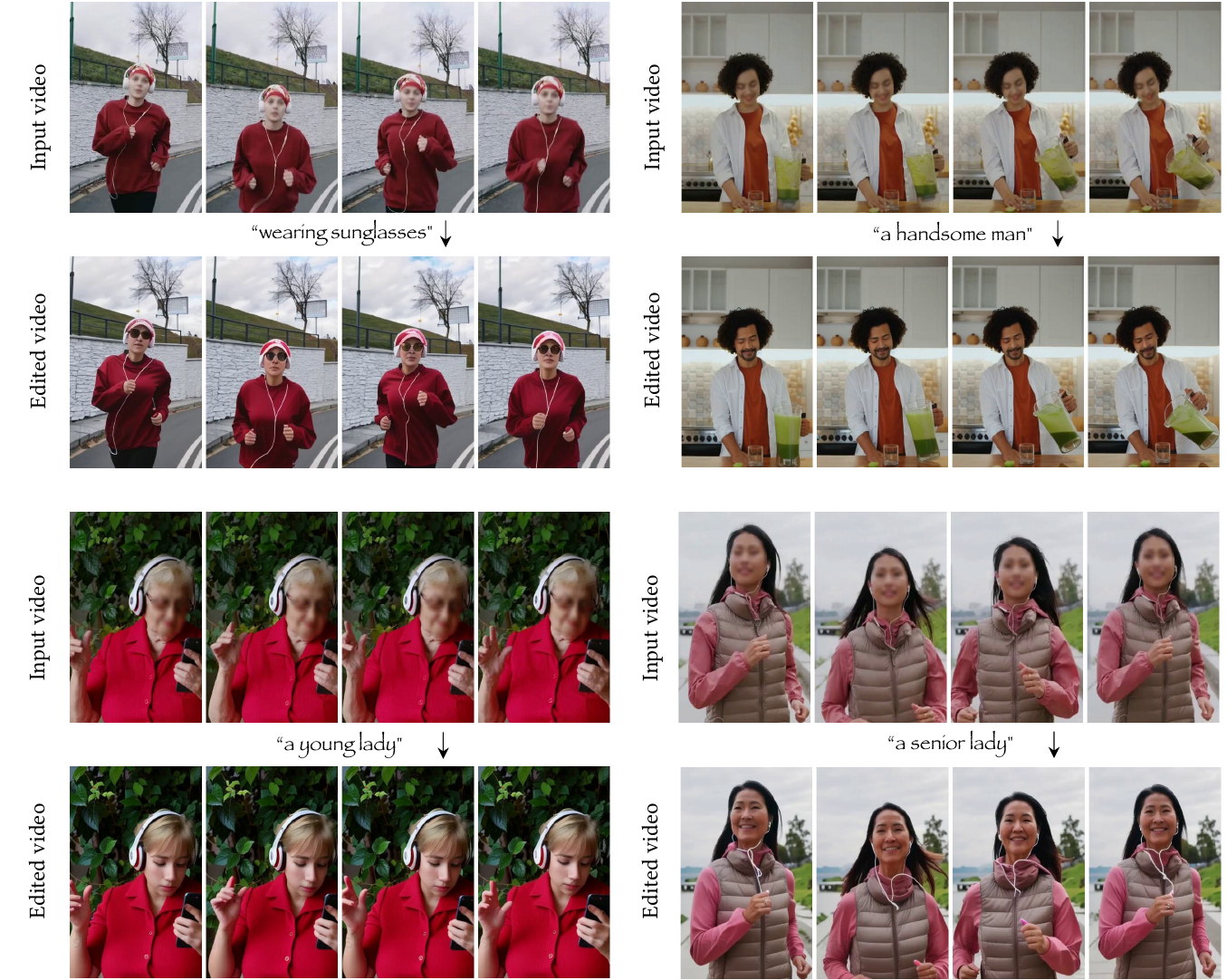}
    \caption{\textbf{Local editing. }
    Given a source video, \modelname enables text-guided local editing (\eg, wearing glasses, or changing gender). \textbf{Faces in source videos are blurred} to protect identities.}
    \label{fig:local_edit}
\end{figure*}

\medskip \noindent \textbf{Local editing}. There are cases when a user only wants to make local modification to the video while leaving other regions untouched (\eg, make the young lady wear glasses as shown in Fig.~\ref{fig:teaser}). 
To achieve this, following SDEdit~\cite{meng2021sdedit}, we first invert the source video via DDIM inversion~\cite{dhariwal2021diffusion,song2020denoising} with a source prompt $\vc_{\rm src}$ describing the original video content. Then, we run the denoising process as usual but with the target prompt $\vc$. Some examples can be found in Fig.~\ref{fig:local_edit}

\begin{figure*}
    \centering
    \includegraphics[width=0.9\textwidth]{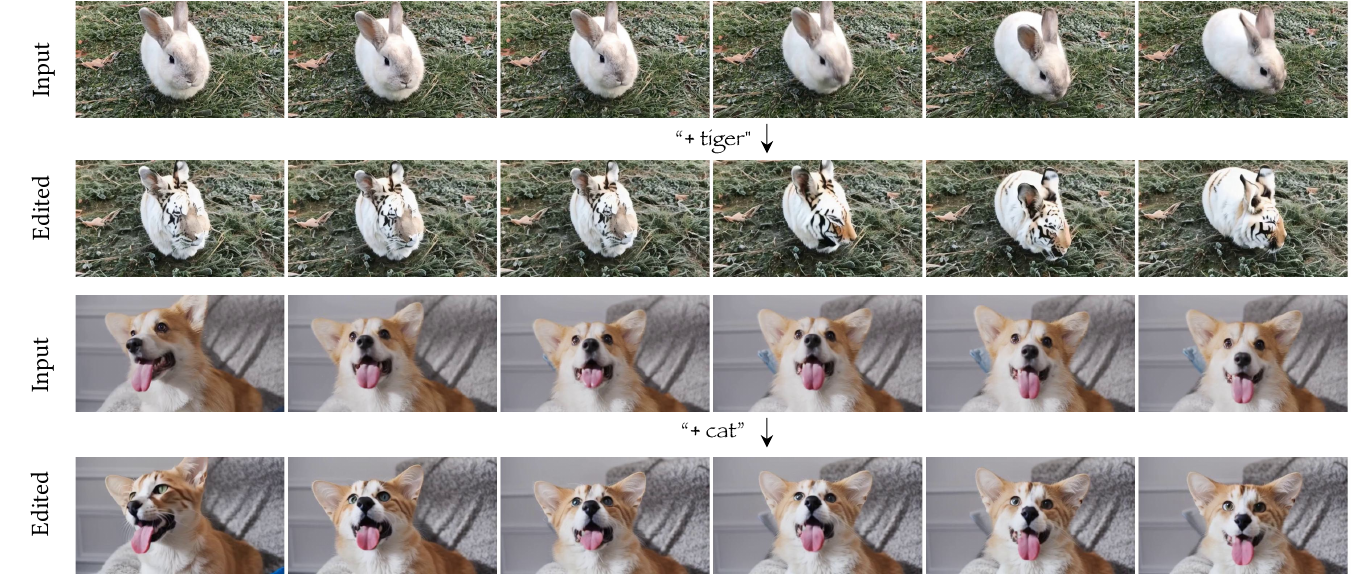}
    \caption{\textbf{Video-MagicMix. }
    Similar to MagicMix~\cite{liew2022magicmix}, \modelname also allows mixing of two different concepts (\eg, ``rabbit'' and ``tiger'') to generate a novel concept (\eg, a rabbit-alike tiger) in the video domain.}
    \label{fig:video_magicmix}
\end{figure*}

\medskip \noindent \textbf{Video-MagicMix. } 
Liew \etal~\cite{liew2022magicmix} previously demonstrated that two different concepts can be mixed to construct a new concept (\eg, ``rabbit'' + ``tiger'' $\rightarrow$ a rabbit-alike tiger). Similarly, we show that MagicMix can be applied to video domain to create a moving rabbit-alike tiger (Fig.~\ref{fig:video_magicmix}).

\begin{figure*}
    \vspace{1em}
    \centering
    \includegraphics[width=\textwidth]{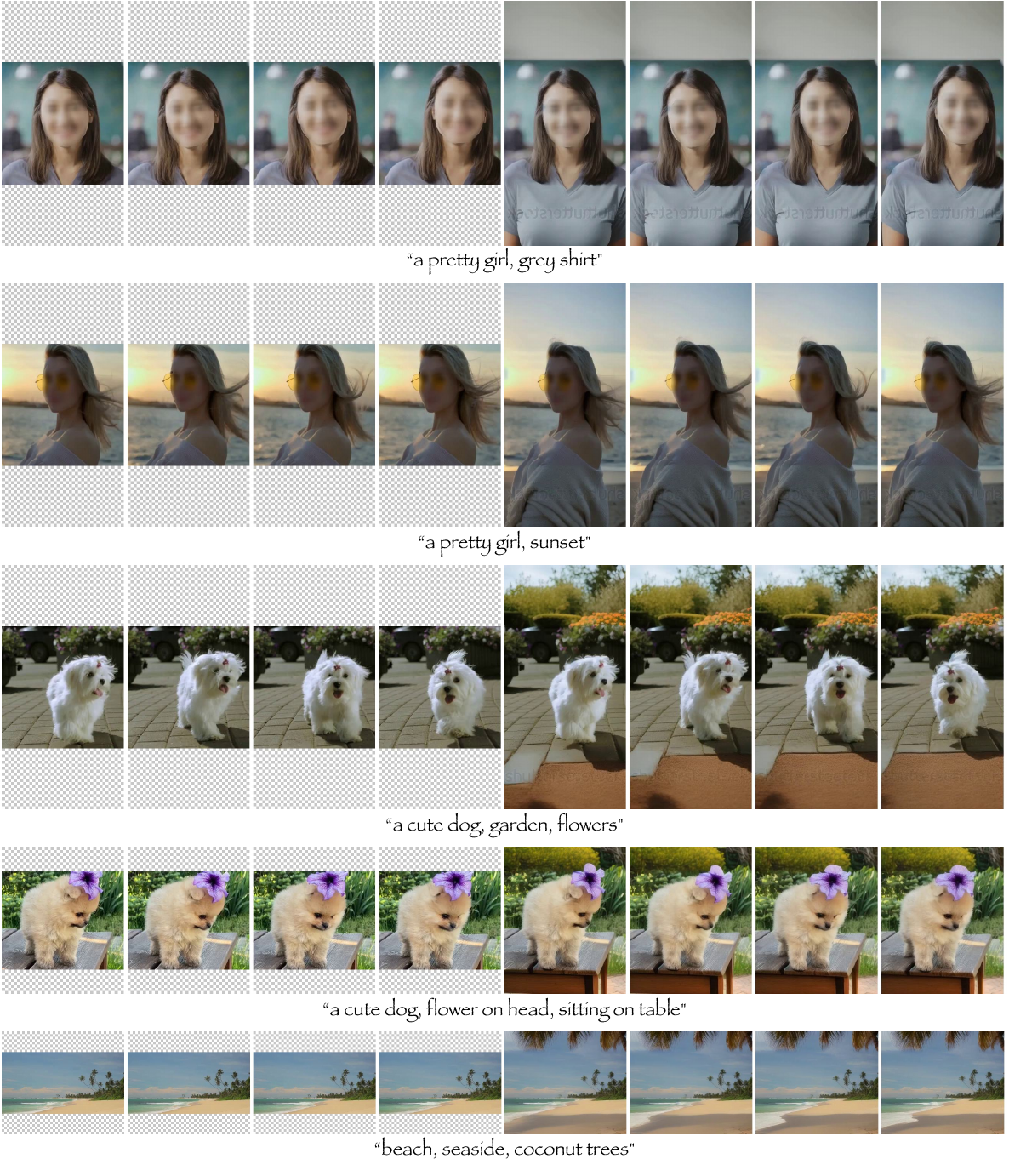}
    \caption{\textbf{Video outpainting. } Our \modelname also supports video outpainting application with various outpainting ratio (see Fig.~\ref{fig:outpaint_different_ratio}).
    \textbf{Faces in source videos are blurred} to protect identities.}
    \label{fig:outpaint}
\end{figure*}

\medskip \noindent \textbf{Video Outpainting. }
We found that \modelname can also be applied for video outpainting task without any re-training. 
Given an input video of spatial size $H\times W$, let the size of the outpainted video be  $h \times w$.
We first invert the source video via DDIM inversion, obtaining a sequence of latents of size $H/8 \times W/8$. Then, we randomly sample $F$ Gaussian noises of size $\times h/8 \times w/8$ and run denoising. 
At each denoising step, we replace the known regions with the inverted latents above to ensure the known areas remain unchanged.
To ensure smooth transition across image borders, we do not replace the known latents for the last few steps.
Some examples of video outpainting are shown in Fig.~\ref{fig:outpaint}.

\begin{figure*}
    \vspace{6em}
    \centering
    \includegraphics[width=\textwidth]{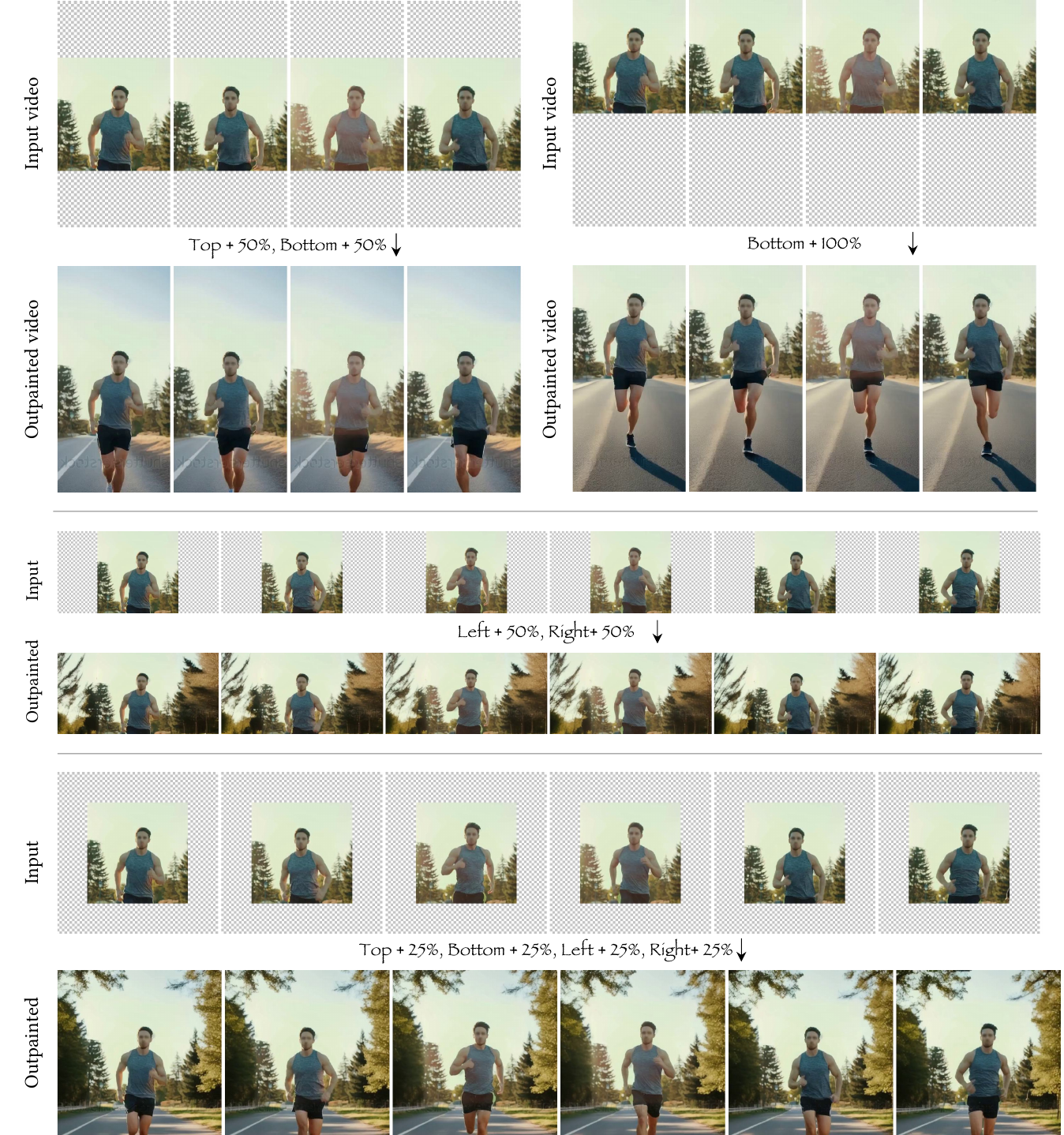}
    \caption{Video outpainting with different ratio.}
    \label{fig:outpaint_different_ratio}
\end{figure*}

\begin{figure}
    \centering
    \includegraphics[width=0.45\textwidth]{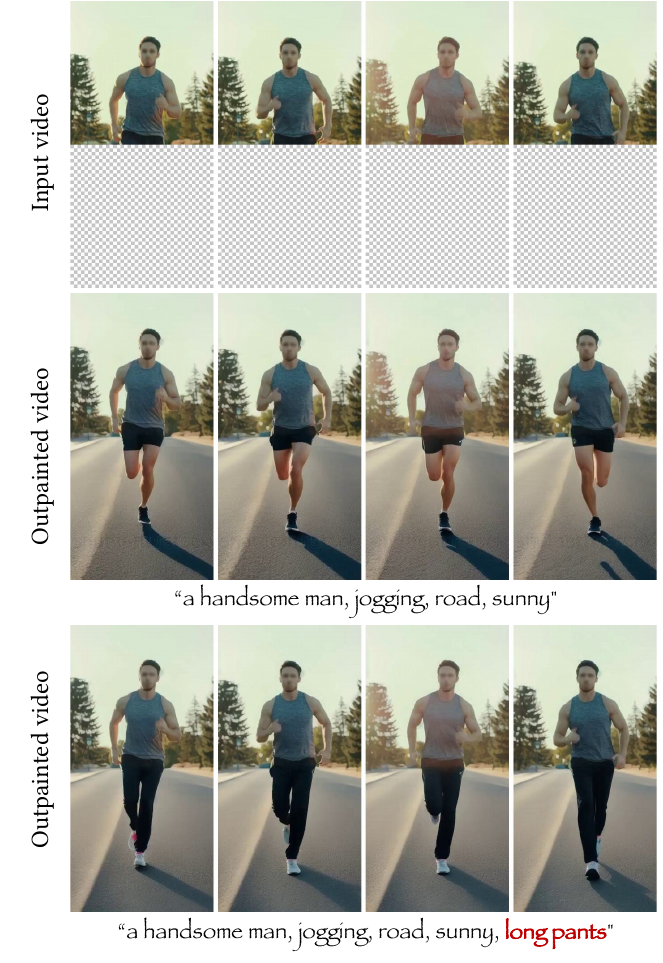}
    \caption{Effects of different prompts on the outpainted video.}
    \label{fig:outpaint_different_prompts}
\end{figure}

In Fig.~\ref{fig:outpaint_different_ratio}, we can also see that the model can handle various ratios, including horizontal, vertical, and even large outpainting ratio (\eg, bottom + 100\%). More interestingly, as shown in Fig.~\ref{fig:outpaint_different_prompts}, our model is also capable to generate different contents by giving different text prompts (\eg, short or long pants), allowing the users to outpaint a video flexibly.

\section{Conclusion}
In this technical report, we present \modelname, a surprisingly simple recipe for effective training of a video editing tool. Our findings show that, high-fidelity and temporally coherent video-to-video translation can be obtained by explicitly disentangling the learning of content, structure and motion signals during training. As a result, \modelname supports a wide variety of downstream editing applications, including video stylization, local editing, video-MagicMix and video outpainting.

{\small
\bibliographystyle{ieee_fullname}
\bibliography{refs}
}

\end{document}